\def\year{2019}\relax
\documentclass[letterpaper]{article} 
\usepackage{aaai19}  
\usepackage{times}  
\usepackage{helvet}  
\usepackage{courier}  
\usepackage{url}  
\usepackage{graphicx}  
\usepackage[utf8]{inputenc}
\usepackage{syntax}
\usepackage{color}
\usepackage{amsthm,amsmath}
\usepackage{listings}
\usepackage{paralist}

\begin{document}
%
\title{HDDL -- A Language to Describe Hierarchical Planning Problems}

\author{D. H\"oller$^{*}$, G. Behnke$^{*}$, P. Bercher$^{*}$, S. Biundo$^{*}$, H. Fiorino$^{\dag}$, D. Pellier$^{\dag}$ and R. Alford$^{\ddag}$\\
$^{*}$Institute of Artificial Intelligence, Ulm University, 89081 Ulm, Germany\\
\{daniel.hoeller, gregor.behnke, pascal.bercher, susanne.biundo\}@uni-ulm.de\\
$^{\dag}$University Grenoble Alpes, LIG, F-38000 Grenoble, France \\
\{humbert.fiorino, damien.pellier\}@imag.fr\\
$^{\ddag}$The MITRE Corporation, McLean, Virginia, USA\\
ralford@mitre.org}
 
\maketitle
\begin{abstract}
The research in hierarchical planning has made considerable progress in the last few years. Many recent systems do not rely on hand-tailored advice anymore to find solutions, but are supposed to be domain-independent systems that come with sophisticated solving techniques. In principle, this development would make the comparison between systems easier (because the domains are not tailored to a single system anymore) and -- much more important -- also the integration into other systems, because the modeling process is less tedious (due to the lack of advice) and there is no (or less) commitment to a certain planning system the model is created for. However, these advantages are destroyed by the lack of a common input language and feature set supported by the different systems. In this paper, we propose an extension to PDDL, the description language used in non-hierarchical planning, to the needs of hierarchical planning systems. We restrict our language to a basic feature set shared by many recent systems, give an extension of PDDL's EBNF syntax definition, and discuss our extensions with respect to several planner-specific input languages from related work. 
\end{abstract}

\newcommand{\mDel}[4]{\ensuremath{\mathit{m\mhyphen{}del}(#1, #2, #3, #4)}}
\newcommand{\mGtDir}[3]{\ensuremath{\mathit{m\mhyphen{}direct}(#1, #2, #3)}} 
\newcommand{\mGtNoop}[2]{\ensuremath{\mathit{mnoop}(#1, #2)}}

\newcommand{\deliver}[2]{\ensuremath{\mathit{deliver}(#1, #2)}}
\newcommand{\getto}[1]{\ensuremath{\mathit{get\mhyphen{}to}(#1)}}
\newcommand{\pickup}[2]{\ensuremath{\mathit{pickup}(#1, #2)}}
\newcommand{\drop}[2]{\ensuremath{\mathit{drop}(#1, #2)}}
\newcommand{\drive}[2]{\ensuremath{\mathit{drive}(#1, #2)}}

\newcommand{\pAt}[2]{\ensuremath{\mathit{at_p}(#1, #2)}}
\newcommand{\tAt}[1]{\ensuremath{\mathit{at_t}(#1)}}
\newcommand{\pIn}[1]{\ensuremath{\mathit{in}(#1)}}
\newcommand{\pRoad}[2]{\ensuremath{\mathit{road}(#1, #2)}}


\newcommand{\PSays}[1] {}
\newcommand{\RSays}[1] {}
\newcommand{\DSays}[1] {}
\newcommand{\GSays}[1] {}

\newcommand{\Shortcite}[1] {\citeauthor{#1} \shortcite{#1}}
\newcommand{\Shortcites}[2] {\citeauthor{#1} \shortcite{#1,#2}}

\newtheorem{Def}{Definition}

\mathchardef\mhyphen="2D

\section{Introduction}
During the past years, much progress has been made in the field of hierarchical planning. Novel systems based on the traditional, search-based techniques have been introduced~\cite{BitMonnotSD16,ShivashankarAA17,Bercher17AdmissibleHTNHeuristic,pandapro18}, but also new techniques like the translation to STRIPS/ADL~\cite{AlfordKN09,Alford2016Bound}, or revisited approaches like the translation to propositional logic~\cite{Behnke2018totSAT,Behnke2018treeSAT,Behnke2019orderchaos,Schreiber2019SAT}. In contrast to earlier systems, all given systems can be considered to be domain-independent, i.e., they do not rely on hand-tailored advice to solve planning problems, but on their solving techniques.

Even though the systems share the basic idea of being \emph{hierarchical} planning approaches, the feature set supported by the different systems is wide-spread: \citeauthor{BitMonnotSD16}~\shortcite{BitMonnotSD16} focuses, e.g., on advanced support for temporal planning, but lack the support for recursion; several systems are restricted to models that do not include partial ordering \cite{AlfordKN09,Behnke2018totSAT,Schreiber2019SAT}; and some, like the one by \citeauthor{ShivashankarAA17}~\shortcite{ShivashankarAA17} even define an entirely new type of hierarchical planning problems.

Even systems restricted to the maybe best-known and most basic hierarchical formalism, called \emph{hierarchical task network} (HTN) planning, do not share a common input language, though the differences between the input languages are sometimes rather subtle, e.g.\ between the formalisms of \citeauthor{Alford2016Bound}~\shortcite{Alford2016Bound} and \citeauthor{Bercher17AdmissibleHTNHeuristic}~\shortcite{Bercher17AdmissibleHTNHeuristic}. To the best of our knowledge, the hierarchical language introduced for the first International Planning Competition (IPC) by \citeauthor{McDermott98PDDL}~\shortcite{McDermott98PDDL} is not supported by any recent system.

The lack of a common language has several consequences for the field.
First, it makes the comparison between the systems tedious due to the translation process. 
%
Second -- and even more important -- it makes the use of hierarchical planning from a practical perspective laborious, because it is not possible to model a problem at hand and try which system performs best on the final model. Selecting the best-fitting system in beforehand (if possible) requires much insights into the systems.

A common description language would make the comparison of the systems easier, it could foster a common set of supported features and result in a common benchmark set the systems are evaluated on.

In this paper, we propose the \emph{hierarchical domain definition language} (HDDL) as common input language for hierarchical planning problems. It is widely based on and fully compatible to the input language of the planners by \citeauthor{Bercher17AdmissibleHTNHeuristic}~\shortcite{Bercher17AdmissibleHTNHeuristic}, \citeauthor{pandapro18}~\shortcite{pandapro18}, \citeauthor{Behnke2018totSAT}~\shortcite{Behnke2018totSAT}, and \citeauthor{Behnke2019orderchaos}~\shortcite{Behnke2019orderchaos}.
We define it as an extension of the STRIPS fragment (language level 1) of the PDDL2.1 definition~\cite{FoxL03}.
To concentrate on a set of features shared by many systems, we restricted the language to basic HTN planning. However, we hope that the given definition is just the starting point for further language extensions like the first PDDL in classical planning.

We start by introducing a lifted HTN planning formalism from the literate, before we introduce our language \emph{by example}. We go through the new language elements, introduce their syntax and meaning, discuss our design choices and the differences to several approaches from the literature, namely PDDL1.2~\cite{McDermott98PDDL}, SHOP(2)~\cite{NauAIKMWY03},
ANML~\cite{smith2008anml}, HPDDL~\cite{Alford2016Bound}, GTOHP~\cite{Ramoul2017GTOHP}, and HTN-PDDL~\cite{gonzalez2009htnpddl}. We then give a full EBNF syntax definition\footnote{When the paper gets accepted, we will provide syntax definitions of our language for the ANTLR and Bison parser generators to support the integration into planning systems.} based on the definition of PDDL2.1 and discuss every extension and change. We conclude with a short outlook.


\section{Lifted HTN Planning}~\label{sec:formalism}
In this section we formally define the problem class HDDL can describe. It is standard hierarchical task network (HTN) planning in line with the text book description by \Shortcite{Ghallab04AutomatedPlanning}. To define the formal framework we extend the formalization of \Shortcites{Alford15TightHTNBounds}{Alford15TightTIHTNBounds}.

Our \emph{lifted} formalism is based upon a quantifier-free first-order predicate logic $\mathcal{L}=(P,T,V,C)$ with the following elements. $P$ is a finite set of \emph{predicate symbols}, each having a finite arity. The arity defines its number of parameter variables (taken from $V$), each having a certain type (defined in $T$). Thus, $T$ is a finite set of \emph{type symbols} as is also known from PDDL. $V$ is a finite set of typed variable symbols to be used by the parameters of the predicates in $P$. $C$ is a finite set of typed constants. They are the syntactic representation of the objects in the real world. Please be aware that a single constant can have several types, e.g.\ \emph{truck} and \emph{vehicle} to support a type hierarchy.

The most basic data structure in HTN planning is a \emph{task network}. Task networks are partially ordered sets of tasks.

In contrast to classical (non-hierarchical) planning, there are two kinds of tasks in HTN planning: primitive and compound ones. Task networks can contain both primitive tasks (also called actions) and compound tasks (also called to be abstract).
%
%
Each task (primitive or compound) is given by its name, followed by a parameter sequence. For instance, a (primitive) task for driving from a source location $\mathit{?ls}$ to a destination location $\mathit{?ld}$ is given by the first-order atom $\mathit{drive(?ls, ?ld)}$. We do not differentiate between the expressions \emph{task} and \emph{task names} -- both are used synonymously.

\begin{Def}[Task Network]
A task network $tn$ over a set of \emph{task names} $X$ (first-order atoms) is a tuple $(I,\mathord{\prec},\alpha,\mathit{VC})$ with the following elements:
\begin{compactenum}
  \item $I$ is a finite (possibly empty) set of \emph{task identifiers}.
  \item $\mathord{\prec}$ is a strict partial order over $I$.
  \item $\alpha : I\rightarrow X$ maps task identifiers to task names.
  \item $\mathit{VC}$ is a set of variable constraints. Each constraint can bind two task parameters to be (non-)equal and it can constrain a task parameter to be (non-)equal to a constant, or to (not) be of a certain type.
\end{compactenum}
\end{Def}

The task identifiers are arbitrary symbols which serve as place holders (or labels) for the actual tasks they represent. We need these identifiers because any task can occur multiple times within the same task network, but the partial order needs to be able to differentiate between them. We call a task network \emph{ground} if all task parameters are bound to (or replaced by) constants from $C$.

Task networks can contain primitive and/or compound tasks. \emph{Primitive tasks} are identical to actions known from classical planning. An \emph{action} $a$ is a tuple $(\emph{name},\emph{pre},\emph{eff})$ with the following elements: \emph{name} is its \emph{task name}, i.e., a first-order atom like $\mathit{drive(?ls, ?ld)}$ consisting of the (actual) name followed by a list of typed parameter variables. \emph{pre} is its \emph{precondition}, a first-order formula over literals over $\mathcal{L}$'s predicates. \emph{eff} is its \emph{effects}, a conjunction of literals over $\mathcal{L}$'s predicates. All variables used in \emph{pre} and \emph{eff} are demanded to be parameters of \emph{name}. We also write \emph{name(a)}, \emph{pre(a)}, and \emph{eff(a)} to refer to these elements. We also require that for each task name \emph{name(a)} there exists only a single action using it as its name (this way, names can be used as unique identifiers).

A \emph{compound task} is simply a task name, i.e., an atom. In contrast to primitive tasks its purpose is not to induce a state transition, but to reference a pre-defined mapping to one or more task networks by which that compound task can then be refined. They do thus not use preconditions or effects. However, there are many hierarchical planning formalisms that do also feature preconditions and/or effects for compound tasks \cite{BercherHBB16}, but they are not within the scope of this paper. The before-mentioned mapping from compound tasks to pre-defined task networks is given by a set of \emph{decomposition methods} $M$. A decomposition method $m\in M$ is a tuple $(c,tn,\mathit{\mathit{VC}})$ consisting of a compound task name $c$, a task network $tn$, and a set of variable constraints $\mathit{VC}$. The variable constraints $\mathit{VC}$ allow to specify (co)designations between the parameters of $c$ and those of the task network $tn$.

\begin{Def}[Planning Domain]
A \emph{planning domain} $\mathcal{D}$ is a tuple $(\mathcal{L},T_P,T_C,M)$ defined as follows.
\begin{compactitem}
 \item $\mathcal{L}$ is the underlying predicate logic.
 \item $T_P$ and $T_C$ are finite sets of primitive and compound tasks, respectively.
 \item $M$ is a finite set of decomposition methods with compound tasks from $T_C$ and task networks over the names $T_P\cup T_C$.
\end{compactitem}
\end{Def}

The domain implicitly defines the set of all states $S$, being defined over all subsets of all ground predicates.

\begin{Def}[Planning Problem]\label{def:problem}
A \emph{planning problem} $\mathcal{P}$ is a tuple $(\mathcal{D},s_I,tn_I,g)$, where:
\begin{compactitem}
  \item $s_I \in S$ is the initial state, a ground conjunction of positive literals over the predicates assuming the closed world assumption.
  \item $tn_I$ is the initial task network that may not necessarily be ground.
  \item $g$ is the goal description, being a first-order formula over the predicates (not necessarily ground).
\end{compactitem}
\end{Def}

HTN planning is \emph{not} about finding courses of action achieving a certain state-based goal definition, so it makes perfect sense to specify no goal formula at all. In fact, goal formulas can trivially be simulated by the task hierarchy \cite{Geier11HybridDecidability}. We added them anyway to be closer to the HDDL specification given later on. Having such a goal formula in the input specification is more convenient in case one actually wants to specify one in the addition to the task hierarchy.

We still need to define the set of possible solutions for a given problem. Informally, solutions are executable, ground, primitive task networks that can be obtained from the problem's initial task network via applying decomposition methods, adding ordering constraints,
and grounding.

Lifted problems are a compact representation of their ground instantiations that are, as in classical planning, up to exponentially smaller \cite{Alford15TightHTNBounds,Alford15TightTIHTNBounds}.
However, we define solutions based on their grounding.
The semantics of such a lifted problem is thus defined in terms of the standard semantics of its ground instantiation. We assume that the reader is familiar with the grounding process and refer to the paper by \Shortcite{Alford15TightHTNBounds} for details about it. To the best of our knowledge there is currently only one publication devoted to grounding in more detail \cite{Ramoul2017GTOHP}\footnote{Please be aware that the procedure described there allows to delete effectless actions \cite{Ramoul2017GTOHP}, which is not allowed in HTN planning. This would, for example, invalidate the compilation process for goal descriptions}. We now give the required definitions based on a \emph{ground} problem and domain. Note that we do not need to represent variable constraints anymore since their constraints are already represented within the groundings.

Given ground problems/models we can now define \emph{executability} of task networks. Let $A$ be the set of ground actions obtained from $T_P$. An action $a\in A$ is called executable in a state $s\in S$ if and only if $s\models pre(a)$. The state transition function $\gamma: S \times A \rightarrow S$ is defined as usual: If $a$ is executable in $s$, then $\gamma(s,a)=(s\setminus del(a))\cup add(a)$, otherwise $\gamma(s,a)$ is undefined. The extension of $\gamma$ to action sequences, $\gamma^*: S \times A^* \rightarrow S$ is defined straightforwardly.

\begin{Def}[Executability]
A task network $tn=(I, \mathord{\prec}, \alpha)$ is called executable if and only if there is a linearization of its task identifiers $i_1,\dots,i_n$, $n=|I|$, such that $\alpha(i_1),\dots,\alpha(i_n)$ is executable in $s_I$.
\end{Def}

\newcommand{\OrdUn}{\ensuremath{\mathord{\prec}}}
The essential means of transforming one task network into another -- to obtain executable task networks -- is decomposition.
\begin{Def}[Decomposition]
Let $m=(c,(I_m,\OrdUn_m,\alpha_m))$ be a decomposition method, $tn_1=(I_1,\OrdUn_1,\alpha_1)$ a task network, and $I_m\cap I_1=\emptyset$ (the latter can be achieved by renaming). Then, $m$ decomposes a task identifier $i\in I_1$ into a task network $tn_2=(I_2,\OrdUn_2,\alpha_2)$ if and only if $\alpha_1(i) = c$ and
	\begin{align*}
        I_2      \; = \; & (I_1\setminus \{i\}) \cup I_m\\
        \OrdUn_2 \; = \; & (\OrdUn_1 \cup \OrdUn_m \cup\\
		                 & \phantom{(}\{(i_1,i_2) \in I_1 \times I_m  \mid (i_1,i) \in \OrdUn_1\} \ \cup \\
		                 & \phantom{(}\{(i_1,i_2) \in I_m  \times I_1 \mid (i,i_2) \in \OrdUn_1\}) \\
		                 & \setminus \{ (i',i'') \in I_1\times I_1\mid i'=i \ or \ i''=i\} \\
		\alpha_2 \; = \; & (\alpha_1 \cup \alpha_m) \setminus \{(i,c)\}
	\end{align*}
\end{Def}

We have now defined all prerequisites required to define the criteria under which a task network can be considered a solution.
%
%

\begin{Def}[Solutions]
Let $\mathcal{P}=(\mathcal{D},s_I,tn_I,g)$ be a planning problem with $\mathcal{D}=(\mathcal{L},T_P,T_C,M)$ and $tn_S = (I_S, \mathord{\prec}_S, \alpha_S)$.
  $tn_S$ is a \emph{solution to an HTN planning problem} $P$ 
  if and only if
  \begin{itemize}
    \item There is a sequence of decompositions from $tn_I$ to $tn=(I,\mathord{\prec},\alpha)$, such that $I=I_S$, $\mathord{\prec}\subseteq\mathord{\prec}_S$, and $\alpha=\alpha_S$
    \item $tn_S$ is primitive and has an executable action linearization leading to a state $s\models g$.
  \end{itemize}
\end{Def}

\section{HDDL by Example}
In this section we explain our extensions to the PDDL definition based on a transport domain. To keep the example simple, the domain includes only a single transporter that has to deliver one or more packages. For each new language element we introduce its syntax and meaning and discuss the way it is modeled in other input languages.

The predicate and type definition is the same as in PDDL:
%
\begin{lstlisting}[escapechar=~]
(define (domain transport)
  (:types location package - object)
  (:predicates
     (road ?l1 ?l2 - location)
     ...)
\end{lstlisting}

\begin{figure}
 \centering
 \scalebox{0.9}{
   \includegraphics{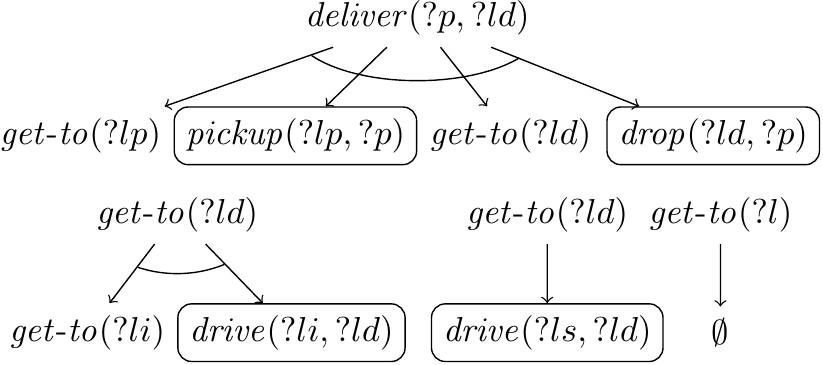}
 }
\caption{The method set of a simple transport domain. Actions are given as boxed nodes, abstract tasks are unboxed. All methods are totally ordered.}\label{fig:methods}
\end{figure}

The full method set of the domain is illustrated in Figure~\ref{fig:methods}. Each method will be discussed in this section.

The domain contains two abstract tasks $\mathit{deliver}$ and $\mathit{get\mhyphen{}to}$.
We propose to include an explicit definition of abstract tasks as it is the case for actions.
HPDDL~\cite{Alford2016Bound} also defines abstract tasks explicitly, albeit with a slightly different syntax.
Both ANML~\cite{smith2008anml} and HTN-PDDL~\cite{gonzalez2009htnpddl} require an explicit declaration of abstract tasks and their parameter types as well, but here the declaration is not separated from other elements of the domain as both declare methods together with their abstract tasks.

Some description languages for HTN problems define abstract tasks only in an implicit way by their use in methods.
This includes the language used by SHOP and SHOP2~\cite{NauAIKMWY03}, PDDL1.2~\cite{McDermott98PDDL}, as well as GTOHP~\cite{Ramoul2017GTOHP}.
SHOP and GTOHP assume that any task that is used in a method, but is not declared to be an action is an abstract task.
In contrast, PDDL1.2 assumes that every task that has no methods is primitive.
This way of implicitly defining the set of compound tasks has also been chosen in some formal definitions of hierarchical problem classes \cite{Alford15TightHTNBounds,Alford15TightTIHTNBounds}.
However, this can be very cumbersome when debugging domains.
If the modeler forgot to define a specific primitive task, the domain will still be valid, as it would be interpreted as an abstract task.

Another problem with such a definition is that the argument types are defined implicitly, namely as those with which the task can be instantiated via any method.
Note that the language of GTOHP~\cite{Ramoul2017GTOHP} further does not allow for using different types (that share a common ancestor in the type hierarchy) to be used for the same task.
For example, there might be different methods for the $\mathit{deliver}$ task, depending on the type of transported package.
$\mathit{deliver}$ might have two methods, one where the first argument is of type $\mathit{regularPackage}$ and one where it is of type $\mathit{valuablePackage}$, the latter requiring an armoured transporter.
We assume that $\mathit{regularPackage}$ and $\mathit{valuablePackage}$ are disjunct types, but have a common super-type $\mathit{package}$, which would be the correct parameter type for $\mathit{deliver}$'s first argument.
If its type is not declared explicitly, the planner can either reject the domain, as GTOHP does, or would have to infer the possible types of the arguments of an abstract task.

Declaring abstract tasks and their parameter types explicitly is also in line with the design choices of PDDL.
Similar to abstract tasks, PDDL could omit the explicit definition of predicates as their types could be inferred from their usages.
This is however discouraged from a modeling point-of-view.

Omitting the distinct definition of tasks and methods would also mean a significant deviation from the contemporary theoretical work on HTN planning. It could also hinder further language extensions like annotating abstract tasks with constraints, e.g.\ preconditions and effects of abstract tasks as used by a couple of systems (see e.g.\ the survey by \citeauthor{BercherHBB16}, \citeyear{BercherHBB16}).

Here is the abstract task definition for the example:
\begin{lstlisting}[firstnumber=last,escapechar=~]
  (:task deliver :parameters (?p - package ~\linebreak~ ?l - location))
  (:task get-to :parameters (?l - location))
\end{lstlisting}

There is only a single method in the model to decompose deliver tasks (given at the top of Figure~\ref{fig:methods}).
It decomposes the task into four ordered sub-tasks: getting to the package, picking it up, getting to its final position, and dropping the package.
The definition in HDDL could look like this:
\begin{lstlisting}[firstnumber=last,escapechar=~]
  (:method m-deliver
    :parameters (?p - package~\linebreak~ ?lp ?ld - location)~\label{l:mexparams}~
    :task (deliver ?p ?ld)~\label{l:mexabstask}~
    :ordered-subtasks (and~\label{l:mexsubtasks}~
      (get-to ?lp)
      (pick-up ?ld ?p)
      (get-to ?ld)
      (drop ?ld ?p)))
\end{lstlisting}

The method definition starts with the name of the method that can e.g.\ be used to describe the decompositions needed to find a solution.
We decided to give the method's parameters explicitly (line~\ref{l:mexparams}).
This allows e.g.\ to restrict the types used in the subtasks to subtypes of the original task parameters.
To be correctly defined, we assume this parameters to be a superset of all parameters used in the entire method definition.
The parameter definition is followed by the specification of the abstract task decomposed by the method as well as its parameters (line~\ref{l:mexabstask}).

The same syntactical structure is used by HPDDL~\cite{Alford2016Bound}.
ANML~\cite{smith2008anml}, PDDL1.2~\cite{McDermott98PDDL}, and HTN-PDDL~\cite{gonzalez2009htnpddl} aggregate all decomposition methods belonging to a single abstract task, which have to be declared as part of the definition of an abstract task.
As such, the variables that are declared as the arguments of an abstract task are automatically variables in a method's task network.

In GTOHP's language, methods don't have names, but are identified via the abstract task they refine.

In SHOP, all variables inside a method are only defined implicitly by their usage as parameters of tasks and predicates inside the method.
For example, the definition of a SHOP method starts with \verb+:method+ followed by an abstract task and its parameters -- which if they are variables are automatically declared as new (untyped) variables.
The same holds for variables that only occur as parameters of a method's subtasks.
GTOHP and HTN-PDDL follow this pattern, but enforce that the parameters of the abstract task are typed, i.e.\ declared explicitly.
Their languages however do not allow to specify the types of variables that occur in the method that are not parameters of the abstract task.
Declaring the variables is, again, in line with the PDDL standard and e.g.\ done the same way in actions. We think it less error-prone. When the modeler explicitly defines the variables and their types, the system can check the compatibility of types and warn the modeler when undeclared variables are used (e.g.\ due to a spelling error).

The subtasks of the method are given afterwards (starting in line~\ref{l:mexsubtasks}).
We decided to have two keywords to start the definition \verb+:ordered-subtasks+ (as given here) and \verb+:subtasks+ (which we will show in the next method).
When the \verb+:ordered-subtasks+ keyword is used, the given list of subtasks is supposed to be totally ordered.
HPDDL~\cite{Alford2016Bound} uses the keyword \verb+:tasks+, which might cause errors if mixed up with the \verb+:task+ keyword.
Since GTOHP~\cite{Ramoul2017GTOHP} does only support totally-ordered HTN planning problems, their language only allows for specifying sequences of actions with the keyword \verb+:expansion+.

In the subtask section, all abstract tasks and actions defined in the domain may be used as subtasks, all variables defined in the method's parameter section may be used as parameters of the subtasks.

The $\mathit{get\mhyphen{}to}$ task form our example domain is again abstract and may be decomposed by using one of the three methods given at the bottom of Figure~\ref{fig:methods}.
We start with the left one that is used when there is no direct road connection. Then the transporter needs to go to the final location $?ld$ via some intermediate location $?li$. Therefore the method decomposes the task into another abstract $\mathit{get\mhyphen{}to}$ task, followed by a $\mathit{drive}$ action with the destination location $?ld$.
\begin{lstlisting}[firstnumber=last,escapechar=~]
  (:method m-drive-to-via
    :parameters (?li ?ld - location)
    :task (get-to  ?ld)
    :subtasks (and~\label{l:mexparord}~
      (t1 (get-to ?li))
      (t2 (drive ?li ?ld)))
    :ordering (and~\label{l:mexordering}~
      (t1 < t2)))
\end{lstlisting}
Line~\ref{l:mexparord} shows the aforementioned \verb+:subtask+ definition that allows for partially ordered tasks.
The task definition contains IDs that can be used to define ordering constraints (line~\ref{l:mexordering}).
They consist of a list of individual ordering constraints between subtasks, identified by their IDs.
However, in the given example the resulting ordering is, again, a total order (and is just defined that way to demonstrate this kind of definition).

HPDDL~\cite{Alford2016Bound} uses the same keyword, but a slightly different syntax so specify ordering constraints.
\citeauthor{Alford2016Bound}'s format omits the \verb+and+ and the \verb+<+ signs.
We would argue that our notation is better readable to humans.
As stated above, GTOHP~\cite{Ramoul2017GTOHP} cannot specify partial orders.
ANML -- as it is primarily designed for temporal domains -- uses a temporal syntax, e.g.\ \verb+end(t1) < start(t2)+.
Lastly, SHOP2 and HTN-PDDL use a different approach to represent the task ordering.
Instead of specifying individual ordering constraints, they require to specify the order as a single expression.
This expression is a nested definition of the ordering, which can only contain two constructors: \verb+ordered+ and \verb+unordered+.
In SHOP2, e.g.\ \verb+((:unordered (t1 t2) t3) t4)+ corresponds to the ordering constraints
\verb+t1 < t2+,
\verb+t2 < t4+, and
\verb+t3 < t4+.
Note that this construction cannot express all possible partially-ordered sets of tasks.
Consider an ordering over five task identifiers \verb+t1+, $\dots$, \verb+t5+, where
\verb+t1 < t4+,
\verb+t2 < t4+,
\verb+t2 < t5+, and
\verb+t3 < t5+.
This ordering cannot be expressed with SHOP's nested ordered/unordered constructs.
PDDL1.2~\cite{McDermott98PDDL} also uses this mode as a default, but does with an additional requirement also allow for an order specification as we and HPDDL do.
Notably PDDL1.2 intertwines the definition of a method's subtasks and the definition of their order.
The syntax of PDDL1.2 to specify the contents of methods and the order of tasks in them is somewhat convoluted and not easily readable.
Thus, we have not adapted their syntax.

A common feature of many HTN planning systems is the possibility of specifying state-based preconditions for methods as supported by the SHOP2 system~\cite{NauAIKMWY03}.
The feature is somehow problematic. First, because it is (at least from our experience) usually used to guide the search and thus often breaks with the philosophy of PDDL to specify a model that does not include advice.
The second problem is the way it is usually realized in the HTN planning systems: The systems introduce a new primitive task that holds the method's preconditions. It is added to the method and placed before all other tasks in the method's subtask network. Consider a totally ordered domain: here, the action is executed directly before the other subtasks of the method and the position where the preconditions are checked is fine. Now consider a partially ordered domain: here, the newly introduced action is not necessarily placed directly before the other subtasks, but we just know that it is placed somewhere before, i.e., the condition did hold at some point before the other tasks are executed, but may have changed meanwhile.
However, though we are aware of these problems, the feature is often used and thus we integrated it and assume the standard semantics as given above.

The preconditions are defined as follows:
\begin{lstlisting}[firstnumber=last,escapechar=~]
  (:method m-already-there
    :parameters (?l - location)
    :task (get-to ?l)
    :precondition (tAt ?l)~\label{l:mexprecondition}~
    :subtasks ())
\end{lstlisting}
Here the method may be applied in a state where the transporter is already located at its destination.
The given method has therefore no subtasks, but still has to assure that the transporter is at its destination.

Method preconditions are typically featured in languages expressing HTNs.
HPDDL~\cite{Alford2016Bound} uses the same syntax we are proposing, while GTOHP~\cite{gonzalez2009htnpddl} uses, as noted above, a separate \verb+:constraints+ section, where the method precondition has to be specified as a \verb+before+ constraint.
This is (presumably) to allow for other state constraints later on.
PDDL1.2~\cite{McDermott98PDDL} also features method preconditions, but here they are specified as part of the task network.
In ANML~\cite{smith2008anml}, there is no explicit means for writing down method preconditions, but they can be encoded into the state constraints allowed by ANML.

There is a strong contrast between what can be expressed in SHOP\footnote{This potentially also applied to HTN-PDDL, as they use a similar syntax. Their description is unfortunately not explicit of the critical point in semantics \cite{gonzalez2009htnpddl}.} and all other HTN formats.
In SHOP, several methods for the same abstract task can be arranged in a single method declaration, each featuring its own method precondition.
For the $i$\textsuperscript{th} method to be usable, it is not sufficient that its precondition is satisfied.
In addition, the preconditions of all previous methods have to be not satisfied as well.
Thus SHOP's method preconditions are in essence a chain of if-else constructs.
This structure can be compiled into several individual methods with preconditions.
In case one of the preconditions contains an existential quantifier (or in SHOP's case a free variable) this leads to universally quantified preconditions in the methods after it.
We never-the-less propose to drop the ability to use such if-else chains, most notably, since none of the newer languages supports it.
Further, this kind of if-else is essentially a means to guide a depth-first search planner in an efficient way.
Thus it does not constitute physics of the domain, but advice to the planner, which should not be part of the domain description language for a domain independent planner.
\RSays{It seems kind of strong to say it does not constitute physics.  It's definitely a constraint on the search space, which differs from other types of advice in SHOP2 (method ordering, for example)}

In addition to method preconditions, HPDDL~\cite{Alford2016Bound} also features method effects, which are modeled after SHOP2's~\cite{NauAIKMWY03} assert and retract functionality.
Method effects are executed in the state in which the method preconditions are evaluated.
As far as we know, their formal semantics is not defined in any publication.
We propose to drop this feature (at least for the given definition that is intended to be the core language), as we argue that it is not commonly used and might be difficult to use for newcomers to HTN planning.
Note that even without method effects in the description language, we can still simulate them with additional actions in the methods' definitions.

Sometimes it might be useful to define constraints in a method, e.g.\ on its variables or sorts. This is demonstrated in the following example where the transporter's source position must be different from its destination.
\begin{lstlisting}[firstnumber=last,escapechar=~]
  (:method m-direct
    :parameters (?ls ?ld - location)
    :task (get-to ?ld)
    :constraints~\label{l:mexconstraint}~
      (not (= ?li ?ld))
    :subtasks (drive ?ls ?ld))
\end{lstlisting}
We are aware that PDDL allows for variable constraints in the preconditions of actions. Due to consistency we also argue to allow this when method preconditions as given above are specified. However, many HTN models are defined without methods that have preconditions and we think it not intuitive to specify a \verb+precondition+ section solely to define variable constraints. Furthermore, we think that other constraints apart from simple variable constraints might be added to the standard.
Therefore we integrated a constraint section to the method definition (line~\ref{l:mexconstraint}f) though our current definition only allows for equality and inequality constraints.

HPDDL~\cite{Alford2016Bound} places the variable constraints of a method into the method's preconditions.
In addition to equality and inequality it features type constraints, where e.g.\ \verb+(valuablePackage ?p)+ is the constraint that \verb+?p+ belongs to the type \verb+valuablePackage+.
GTOHP~\cite{Ramoul2017GTOHP} allows for equality and inequality constraints that are also within the \verb+:constraints+ section, but are located in a separate \verb+before+ block.
In SHOP's syntax, variable constraints have to be compiled into method preconditions referring to predicates for the individual types and an explicitly declared \verb+equals+ predicate.
ANML also allows for variable constraints that can be declared freely anywhere inside a method.

We left the action definition unchanged compared to the PDDL standard we build on. Therefore we included only the following action into our example.
\begin{lstlisting}[firstnumber=last,escapechar=~]
  (:action drive
    :parameters (?l1 ?l2 - location)
    :precondition (and
      (tAt ?l1)
      (road ?l1 ?l2))
    :effect (and
      (not (tAt ?l1))
      (tAt ?l2)))
  ...)
\end{lstlisting}

The problem file is slightly adapted to represent the additional elements necessary for HTN panning.
\begin{lstlisting}[escapechar=~]
(define (problem p)
 (:domain transport)
 (:objects
  city-loc-0 city-loc-1 city-loc-2 - location
  package-0 package-1 - package)
 (:htn
  :tasks (and
   (deliver package-0 city-loc-0)
   (deliver package-1 city-loc-2))
  :ordering ()
  :constraints ())
 (:init
  (road city-loc-0 city-loc-1)
  (road city-loc-1 city-loc-0)
  (road city-loc-1 city-loc-2)
  (road city-loc-2 city-loc-1)
  (at package-0 city-loc-1)
  (at package-1 city-loc-1)))
\end{lstlisting}
We consider the term the section starts with as specification of the problem class. In this example, it starts with \verb+:htn+ to define a standard HTN planning problem. However, extensions to the language might add new classes. An example for such another class may be HTN planning with task insertion, where the planner is allowed to insert tasks apart from the hierarchy. These problems are syntactically equivalent to standard HTN planning problems, so we need the given flag to specify the problem class.

The definition of the initial task network is nested in this section.
It has the same form as the methods' subtask networks.
The other description languages for HTN planning also allow for a similar definition of the initial plan.
Again, all of them use a slightly different syntax to describe them.

In the given example, the planning process is started with two $\mathit{deliver}$ tasks, one for each package. These initial tasks are not ordered with respect to each other, i.e.\ their subtasks may be executed interleaved.

In the original PDDL standard, the domain designer has to specify a state-based goal.
HTN planning problems do not require such a goal and thus often do not specify one.
Therefore we made its definition optional.

\section{Full Syntax Definition}
\newcommand{\specReq}[1]{\ensuremath{\mathtt{^{#1}}}}

%
%
We wanted to define our syntax as close as possible to the STRIPS part (i.e.\ language level 1) of the PDDL 2.1 language definition of~\citeauthor{FoxL03}~\shortcite{FoxL03}. Wide parts of the following definition are \textbf{identical} to their definition. Changes and extensions are discussed in the following.

The domain definition has been extended by definitions for compound tasks (line~\ref{l:compTask}) and methods (line~\ref{l:methods}).
\begin{lstlisting}[escapechar=~]
<domain> ::= (define (domain <name>)
    [<require-def>]
    [<types-def>]~\specReq{:typing}~
    [<constants-def>]
    [<predicates-def>]
    <comp-task-def>*~\label{l:compTask}~
    <method-def>*~\label{l:methods}~
    <action-def>*)
\end{lstlisting}

%
%
\noindent
The definition of the basic domain elements is nearly unchanged.
\begin{lstlisting}[firstnumber=last, escapechar=~]
<require-def> ::= ~\linebreak~(:requirements <require-key>+)
<require-key> ::= ~\dots~
<types-def> ::= (:types <types>+)~\label{l:typesDef}~
<types> ::= <typed list (name)> ~\linebreak~| <base-type>
<base-type> ::= <name>~\label{l:typesDef2}~
<constants-def> ::= ~\linebreak~(:constants <typed list (name)>)
<predicates-def> ::= ~\linebreak~(:predicates <atomic-formula-skeleton>+)
<atomic-formula-skeleton> ::= ~\linebreak~(<predicate> <typed list (variable)>)
<predicate> ::= <name>
<variable> ::= ?<name>
<typed list (x)> ::= x+ - <type> ~\linebreak~[<typed list (x)>]~\label{l:typedlist}~
<primitive-type> ::= <name>
<type> ::= (either <primitive-type>+)
<type> ::= <primitive-type>
\end{lstlisting}

The only change concerns the definition of \verb+<types-def>+ (lines~\ref{l:typesDef} and \ref{l:typesDef2}) in combination with the definition of \verb+<typed list (name)>+ (line~\ref{l:typedlist}). In the PDDL2.1 standard, this can be realized by a list of names, e.g.\ in an untyped way. Our intention was to enforce a typed model and therefore allow for untyped elements only in the type definition. There, it is necessary to define the base type(s). In every other definition that includes \verb+<typed list (name)>+ (e.g.\ parameter and constant definitions), we wanted to enforce a typed list.

%
%
Abstract tasks are defined similar to actions.
\begin{lstlisting}[firstnumber=last, escapechar=~]
<comp-task-def> ::= (:task <task-def>)
<task-def> ::= <task-symbol> ~\linebreak~:parameters (<typed list (variable)>)
<task-symbol> ::= <name>
\end{lstlisting}

%
%
In a standard HTN setting, methods consist of a parameter list (line~\ref{l:mparams}), the abstract task they decompose (line~\ref{l:mabstask}), and the resulting task network (line~\ref{l:msubtasks}). The parameters of a method are supposed to include all parameters of the abstract task that it decomposes and those of the tasks in its network of subtasks. The separate definition of method parameters enables e.g.\ the restriction of the abstract task's parameters to subtypes of their original definition.

By setting the \verb+:htn-method-prec+ requirement, one might use method preconditions as discussed above (line~\ref{l:mprec}).

\begin{lstlisting}[firstnumber=last, escapechar=~]
<method-def> ::= (:method <name>
    :parameters (<typed list (variable)>)~\label{l:mparams}~
    :task (<task-symbol> <term>*)~\label{l:mabstask}~
    [:precondition <gd>]~\specReq{:htn-method-prec}~~\label{l:mprec}~
    <tasknetwork-def>~\label{l:msubtasks}~)
\end{lstlisting}

%
%
The definition of task networks is used in method definitions as well as in the problem definition to define the initial task network.
It contains the definition of sub-tasks (line~\ref{l:tnsubtasks}), ordering constraints (line~\ref{l:tnordering}), and variable constraints (line~\ref{l:tnconstraints}) between any method parameters.

When the key \verb+:ordered-subtasks+ is used, the network is regarded to be totally ordered. In the other cases, ordering relations may be defined explicitly. This is done by including ids into the task definition that can then be referenced in the ordering definition.
\begin{lstlisting}[firstnumber=last, escapechar=~]
<tasknetwork-def> ::=
    [:[ordered-][sub]tasks ~\linebreak~<subtask-defs>]~\label{l:tnsubtasks}~
    [:order[ing] <ordering-defs>]~\label{l:tnordering}~
    [:constraints <constraint-defs>]~\label{l:tnconstraints}~
\end{lstlisting}
We use the same syntax definition for method subnetworks and the initial task network. Here, the keyword \verb+subtasks+ would seem odd. Therefore the syntax also allows for the keys \verb+tasks+ and \verb+ordered-tasks+ (line~\ref{l:tnsubtasks}) that are supported to be used in the initial task network. 

%
%
The subtask definition may contain one or more subtasks. A single task consists of a task symbol and a list of parameters. In case of a method's subnetwork, these parameters have to be included in the method's parameters, in case of the initial task network, they have to be defined as constants in $s_0$ or in a dedicated parameter list (see definition of the initial task network, line~\ref{l:tniparams}). The tasks may start with an id that can be used to define ordering constraints.
\begin{lstlisting}[firstnumber=last, escapechar=~]
<subtask-defs> ::= () | <subtask-def> ~\linebreak~| (and <subtask-def>+)
<subtask-def> ::= (<task-symbol> <term>*) ~\linebreak~| (<subtask-id> (<task-symbol> <term>*))
<subtask-id> ::= <name>
\end{lstlisting}

%
%
The ordering constraints are defined via the task ids. They have to induce a partial order.
\begin{lstlisting}[firstnumber=last, escapechar=~]
<ordering-defs> ::= () | <ordering-def> ~\linebreak~| (and <ordering-def>+)
<ordering-def> ::= ~\linebreak~(<subtask-id> "<" <subtask-id>)
\end{lstlisting}

%
%
So far we only included variable constraints into the constant section, but the definition might be extended in further language levels, of course.
\begin{lstlisting}[firstnumber=last, escapechar=~]
<constraint-defs> ::= () | <constraint-def> | (and <constraint-def>+)
<constraint-def> ::= () ~\linebreak~| (not (= <term> <term>)) ~\linebreak~| (= <term> <term>)
\end{lstlisting}

%
%
The original action definition of PDDL has been split to reuse its body in the task definition.
\begin{lstlisting}[firstnumber=last, escapechar=~]
<action-def> ::= (:action <task-def>
    [:precondition <gd>]
    [:effects <effect>])
\end{lstlisting}

%
We restricted the definition of preconditions and effects to level 1, i.e.\ the STRIPS part of the overall language.
\begin{lstlisting}[firstnumber=last, escapechar=~]
<gd> ::= ()
<gd> ::= <atomic formula (term)>
<gd> ::=~\specReq{:negative\mhyphen{}preconditions}~ <literal (term)>
<gd> ::= (and <gd>*)
<gd> ::=~\specReq{:disjunctive\mhyphen{}preconditions}~ (or <gd>*)
<gd> ::=~\specReq{:disjunctive\mhyphen{}preconditions}~ (not <gd>)
<gd> ::=~\specReq{:disjunctive\mhyphen{}preconditions}~ (imply <gd> <gd>)
<gd> ::=~\specReq{:existential\mhyphen{}preconditions}~ ~\linebreak~(exists (<typed list (variable)>*) <gd>)
<gd> ::=~\specReq{:universal\mhyphen{}preconditions}~ ~\linebreak~(forall (<typed list (variable)>*) <gd>)
<gd> ::= (= <term> <term>)
\end{lstlisting}

\begin{lstlisting}[firstnumber=last, escapechar=~]
<literal (t)> ::= <atomic formula(t)>
<literal (t)> ::= (not <atomic formula(t)>)
<atomic formula(t)> ::= (<predicate> t*)
\end{lstlisting}

\begin{lstlisting}[firstnumber=last, escapechar=~]
<term> ::= <name>
<term> ::= <variable>
\end{lstlisting}

%
%
\begin{lstlisting}[firstnumber=last, escapechar=~]
<effect> ::= ()
<effect> ::= (and <c-effect>*)
<effect> ::= <c-effect>
<c-effect> ::=~\specReq{:conditional\mhyphen{}effects}~ ~\linebreak~(forall (<variable>*) <effect>)
<c-effect> ::=~\specReq{:conditional\mhyphen{}effects}~ ~\linebreak~(when <gd> <cond-effect>)
<c-effect> ::= <p-effect>
<p-effect> ::= (not <atomic formula(term)>)
<p-effect> ::= <atomic formula(term)>
<cond-effect> ::= (and <p-effect>*)
<cond-effect> ::= <p-effect>
\end{lstlisting}

%
%
The problem definition includes as additional element the initial task network (line~\ref{l:tnihtn}). Since a state-based goal definition is often not included in HTN planning, we made the goal definition optional (line~\ref{l:goal}).
\begin{lstlisting}[firstnumber=last, escapechar=~]
<problem> ::= (define (problem <name>)
    (:domain <name>)
    [<require-def>]
    [<p-object-declaration>]
    [<p-htn>]~\label{l:tnihtn}~
    <p-init>
    [<p-goal>])~\label{l:goal}~
\end{lstlisting}

\begin{lstlisting}[firstnumber=last, escapechar=~]
<p-object-declaration> ::= ~\linebreak~(:objects <typed list (name)>)
<p-init> ::= (:init <init-el>*)
<init-el> ::= <literal (name)>
<p-goal> ::= (:goal <gd>)
\end{lstlisting}

The initial task network contains the definition of the problem class (line~\ref{l:tniclass}). In this first definition we only included standard HTN planning, but we integrated this definition to allow for other classes, such as, e.g.\ HTN planning with task insertion. 
\begin{lstlisting}[firstnumber=last, escapechar=~]
<p-htn> ::= (<p-class>~\label{l:tniclass}~
    [:parameters (<typed list (variable)>)] ~\label{l:tniparams}~
    <tasknetwork-def>)~\label{l:tnitasks}~
<p-class> ::= :htn~\specReq{:htn}~
\end{lstlisting} 

Our overall definition includes two new requirement flags:
\begin{itemize}
 \item \verb+:htn+ requires the applied system needs to support HTN planning at all, so this can be seen as the basic requirement for the language defined here.
 \item \verb+:htn-method-prec+ requires the applied system needs to support method preconditions
\end{itemize}

\section{Discussion}
We consider the language proposed in this paper as a first step towards a standardized language for hierarchical planning problems and hope that it helps to find a minimal set of features supported by the diverse systems. However, this basic feature set as well as many design options are still open and have to be discussed in the research community.

First of all, we think it is important to remain as close as possible to the PDDL and to reuse its features to allow domain modelers to create both hierarchical and non-hierarchical problems with minimal learning effort. Language compactness is another important feature to facilitate the adoption of this language. Then, we must decide which features have to be at the core of the language, and which ones are secondary and possibly could be ignored.

A feature that was present in the early HTN formalisms (see e.g.\ the formalism used by \citeauthor{Erol96HTNPlanning}, \citeyear{Erol96HTNPlanning}) is the possibility to define more elaborated constraints in task networks. Recent work in hierarchical planning was not based on such a rich definition language, but on rather minimalistic formalisms like the one introduced by \citeauthor{Geier11HybridDecidability}, \shortcite{Geier11HybridDecidability}. In this first definition we only included the very basic constraints: ordering constraints, variable constraints, and method preconditions.
However, we think that a constraint set as given in PDDL3 might be a nice extension beneficial for domain designers.

When the community wants to foster application in real world domains, it may be necessary to integrate support for numbers and  time into the planning systems. Since our definition builds upon the PDDL2.1, at least the extension of the syntax in that direction could easily be made.

Other extensions that might be integrated are e.g.\ preconditions and effects of abstract tasks (see \citeauthor{BercherHBB16}~\shortcite{BercherHBB16} for an overview of that feature) or the ability to decompose not only tasks, but also goals (as e.g.\ done by \citeauthor{ShivashankarKNA12}, \citeyear{ShivashankarKNA12}) that even has been combined with task decomposition~\cite{AlfordSRFA16}.

\section{Conclusion}
In this paper we propose a common description language for hierarchical planning problems. We argue that the core feature set underlying many hierarchical planning systems from the past years is that of HTN planning and introduced its elements as an extension of PDDL, the description language commonly used in non-hierarchical planning. We defined the language in a way that can easily extended by further features as has been done in PDDL.
We introduced our novel language elements ``by example'' and discussed our design choices, the syntax used in related work, and the proposed meaning. We gave a full syntax definition afterwards and discussed the extensions and changes to the PDDL standard.
We hope that a common input language may foster the cooperation between groups working in hierarchical planning, the comparison of different hierarchical planning systems and the application on real problems, because it enables an easy exchange of the planning system used for a given problem.

%

\bibliographystyle{aaai}
\bibliography{literature}

\end{document}